\documentclass{llncs}

\usepackage{amssymb}
\usepackage{graphicx}

\usepackage{url}
\urldef{\mailsa}\path|{dce.sukriti, vagisha.nda}@gmail.com|    
\newcommand{\keywords}[1]{\par\addvspace\baselineskip
\noindent\keywordname\enspace\ignorespaces#1}

\begin{document}

\mainmatter  

\title{Extractive Summarization using Deep Learning}

\titlerunning{Extractive Summarization using Deep Learning}

%
%
\author{Sukriti Verma \and Vagisha Nidhi}

\authorrunning{Sukriti Verma et al.}

\institute{Delhi Technological University\\
Shahbad Daulatpur, Main Bawana Road,\\ 
Delhi-110042, India\\
\mailsa\\
\url{http://dtu.ac.in/}}

%
%

\toctitle{Extractive Summarization using Deep Learning}
\tocauthor{Sukriti Verma and Vagisha Nidhi}
\maketitle

\begin{abstract}
\emph{This paper proposes a text summarization approach for factual reports using a deep learning model. This approach consists of three phases: feature extraction, feature enhancement, and summary generation, which work together to assimilate core information and generate a coherent, understandable summary. We are exploring various features to improve the set of sentences selected for the summary, and are using a Restricted Boltzmann Machine to enhance and abstract those features to improve resultant accuracy without losing any important information. The sentences are scored based on those enhanced features and an extractive summary is constructed. Experimentation carried out on several articles demonstrates the effectiveness of the proposed approach.}
\end{abstract}

\section{Introduction}

A summary can be defined as a text produced from one or more texts, containing a significant portion of the information from the original text(s), and that is no longer than half of the original text(s) \cite{Hovy}. According to \cite{Mani}, text summarization is the process of distilling the most important information from a source (or sources) to produce an abridged version for a particular user and task(s). When this is done by means of a computer, i.e. automatically, we call it Automatic Text Summarization. This process can be seen as a form of  compression and it necessarily suffers from information loss but it is essential to tackle the information overload due to abundance of textual material available on the Internet.

Text Summarization can be classified into extractive summarization and abstractive summarization based on the summary generated. Extractive summarization is creating a summary based on strictly what you get in the original text. Abstractive summarization mimics the process of paraphrasing a text. Text(s) summarized using this technique looks more human-like and produces condensed summaries. These techniques are much harder to implement than the extractive summarization techniques.

In this paper, we follow the extractive methodology to develop techniques for summarization of factual reports or descriptions. We have developed an approach for single-document summarization using deep learning. So this paper intends to propose an approach by referencing the architecture of the human brain. It is broken down into three phases: feature extraction \cite{Chuang}, feature enhancement, and summary generation based on values of those features. Since it can be very difficult to construct high-level, abstract features from raw data, we use deep learning in the second phase to build complex features out of simpler features extracted in the first phase. These extracted features depend highly on how factual the given document is.
In the end, we have run the proposed algorithm on several factual reports to evaluate and demonstrate the effectiveness of the proposed approach based on the measures such as Recall, Precision, and F-measure.

\section{Related Works}

Most early work on text summarization was focused on technical documents and early studies on summarization aimed at summarizing from pre-given documents without any other requirements, which is usually known as generic summarization \cite{Berger}. Luhn \cite{Luhn} proposed that the frequency of a particular word in an article provides a useful measure of its significance. A number of key ideas, such as stemming and stop word filtering, were put forward in this paper that have now been understood as universal preprocessing steps to text analysis. Baxendale \cite{Baxendale} examined 200 paragraphs and found that in 85\% of the paragraphs, the topic sentence came as the first one and in 7\% of the time, it was the last sentence. This positional feature has been used in many complex machine learning based systems since. Edmundson \cite{Edmundson} focused his work around the importance of word frequency and positional importance as features. Two other features were also used: cue words, and the skeleton structure of the document. Weights were associated with these features manually and finally sentences were scored. During evaluation, it was found that around 44\% of the system generated summaries matched the target summaries written manually by humans.

Upcoming researchers in text summarization have approached it problem from many aspects such as natural language processing \cite{Zhang}, statistical modelling \cite{Darling} and machine learning. While initially most machine learning systems assumed feature independence and relied on naive-Bayes methods, other recent ones have shifted focus to selection of appropriate features and learning algorithms that make no independence assumptions. Other significant approaches involved Hidden Markov Models and log-linear models to improve extractive summarization. More recent papers, in contrast, used neural networks towards this goal.

Text Summarization can be done for one document, known as single-document summarization \cite{Wan}, or for multiple documents, known as multi-document summarization \cite{Shen}. On basis of the writing style of the final summary generated, text summarization techniques can be divided into extractive methodology and abstractive methodology \cite{Wong}. The objective of generating summaries via the extractive approach is choosing certain appropriate sentences as per the requirement of a user. Due to the idiosyncrasies of human-invented languages and grammar, extractive approaches, which select a subset of sentences from the input documents to form a summary instead of paraphrasing like a human \cite{Chen}, are the mainstream in the area. 

Almost all extractive summarization methods have three main obstacles. The first obstacle is the ranking problem i.e. how you rank words, phrases and/or sentences. The second obstacle is the selection problem i.e. how to select a subset of those ranked units \cite{Jin}. The third obstacle is the coherence problem i.e. how to ensure that the selected units form an understandable summary rather than being a set of disconnected words, phrases and/or sentences. Algorithms that determine the relevance of a textual unit, that is words, phrases and/or sentences, with respect to the requirement of the user are used to solve the ranking problem. The selection and coherence problems are solved by methods that improve diversity, minimize redundancy and pick up phrases and/or sentences that are somewhat similar so that more relevant information can be covered by the summary in lesser words and the summary is coherent. Our approach solves the ranking problem by learning a certain set of features for each sentence. On the basis of these features, a score is calculated for each sentence and sentences are arranged in decreasing order of their scores \cite{2016}. Even with a list of ranked sentences, it is not a trivial problem to select a subset of sentences for a coherent summary which includes diverse information, minimizes redundancy and is within a word limit. Our approach solves this problem as follows. The most relevant sentence is the first sentence in this sorted list and is chosen as part of the subset of sentences which will form the summary. Then the next sentence selected is a sentence having highest Jaccard similarity with the first sentence and is picked from the top half of the list. This process is recursively and incrementally repeated to select more sentences until limit is reached.

\section{Proposed Approach}
\subsection{Preprocessing}
Preprocessing is crucial when it comes to processing text. Ambiguities can be caused by various verb forms of a single word, different accepted spellings of a certain word, plural and singular terms of the same things. Moreover, words like a, an, the, is, of etc. are known as stop words. These are certain high frequency words that do not carry any information and don’t serve any purpose towards our goal of summarization. In this phase we do:

\begin{enumerate}

\item \textbf{Document Segmentation:} The text is divided into paragraphs so as to keep a track of which paragraph each sentence belongs to and what is the position of a sentence in its respective paragraph.

\item \textbf{Paragraph Segmentation:} The paragraphs are further divided into sentences.

\item \textbf{Word Normalization:} Each sentence is broken down into words and the words are normalized. Normalization involves lemmatization and results in all words being in one common verb form, crudely stemmed down to their roots with all ambiguities removed. For this purpose, we use Porters algorithm.

\item \textbf{Stop Word Filtering:} Each token is analyzed to remove high frequency stop words.

\item \textbf{PoS Tagging:} Remaining tokens are Part-of-Speech tagged into verb, noun, adjective etc. using the PoS Tagging module supplied by NLTK \cite{nltk}. 

\end{enumerate}

\subsection{Feature Extraction}
Once the complexity has been reduced and ambiguities have been removed, the document is structured into a sentence-feature matrix. A feature vector is extracted for each sentence. These feature vectors make up the matrix. We have experimented with various features. The combination of the following 9 sentence features has turned out most suitable to summarize factual reports. These computations are done on the text obtained after the preprocessing phase:

\begin{enumerate}

\item \textbf{Number of thematic words:} The 10 most frequently occurring words of the text are found. These are thematic words. For each sentence, the ratio of no. of thematic words to total words is calculated. 
\begin{equation}
Sentence\_Thematic = \frac{No.\ of\ thematic\ words}{Total\ words} 
\end{equation}

\item \textbf{Sentence position:} This feature is calculated as follows.\\

\begin{equation}
Sentence\_Position = \left\{\begin{array}{l l} 
1, \hbox{if its the first or last sentence of the text}\\ 
cos((SenPos-min)((1/max)-min)), \hbox{otherwise}\\ \end{array} \right.
\end{equation}
where,	SenPos = position of sentence in the text\\
        min = th x N\\
		max = th x 2 x N\\

N is total number of sentences in document\\
th is threshold calculated as 0.2 x N

By this, we get a high feature value towards the beginning and ending of the document, and a progressively decremented value towards the middle. 

\item \textbf{Sentence length:} This feature is used to exclude sentences that are too short as those sentences will not be able to convey much information. 

\begin{equation}
Sentence\_Length = \left\{\begin{array}{l l} 
0, \hbox{if number of words is less than 3}\\ 
No.\ of\ words\ in\ the\ sentence, \hbox{otherwise}\\ \end{array} \right.
\end{equation}

\item \textbf{Sentence position relative to paragraph:} This comes directly from the observation that at the start of each paragraph, a new discussion is begun and at the end of each paragraph, we have a conclusive closing.

\begin{equation}
Position\_In\_Para = \left\{\begin{array}{l l} 
1, \hbox{if it is the first or last sentence of a paragraph}\\ 
0, \hbox{otherwise}\\ \end{array} \right.
\end{equation}

\item \textbf{Number of proper nouns:} This feature is used to give importance to sentences having a substantial number of proper nouns. Here, we count the total number of words that have been PoS tagged as proper nouns for each sentence.

\item \textbf{Number of numerals:} Since figures are always crucial to presenting facts, this feature gives importance to sentences having certain figures. For each sentence we calculate the ratio of numerals to total number of words in the sentence. 

\begin{equation}
Sentence\_Numerals = \frac{No.\ of\ numerals}{Total\ words}
\end{equation}

\item \textbf{Number of named entities:} Here, we count the total number of named entities in each sentence. Sentences having references to named entities like a company, a group of people etc. are often quite important to make any sense of a factual report.

\item \textbf{Term Frequency-Inverse Sentence Frequency (TF – ISF):} Since we are working with a single document, we have taken TF-ISF feature into account rather than TF-IDF. Frequency of each word in a particular sentence is multiplied by the total number of occurrences of that word in all the other sentences. We calculate this product and add it over all words.

\begin{equation}
TF-ISF = \frac {log(\sum_{all\ words} TF*ISF)}{Total\ words}
\end{equation}

\item \textbf{Sentence to Centroid similarity:} Sentence having the highest TF-ISF score is considered as the centroid sentence. Then, we calculate cosine similarity of each sentence with that centroid sentence. 

\begin{equation}
Sentence\_Similarity = cosine\_sim(sentence, centroid) 
\end{equation}

\end{enumerate}

At the end of this phase, we have a sentence-feature matrix.

\subsection{Feature Enhancement}
The sentence-feature matrix has been generated with each sentence having 9 feature vector values. After this, recalculation is done on this matrix to enhance and abstract the feature vectors, so as to build complex features out of simple ones.
This step improves the quality of the summary.\\

\begin{figure}
\centering

\includegraphics[height=4.0cm]{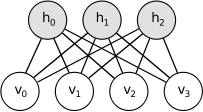}

\caption{A Restricted Boltzmann Machine \cite{dl}}

\label{fig:rbm}

\end{figure}

To enhance and abstract, the sentence-feature matrix is given as input to a Restricted Boltzmann Machine (RBM) which has one hidden layer and one visible layer. A single hidden layers will suffice for the learning process based on the size of our training data. The RBM that we are using has 9 perceptrons in each layer with a learning rate of 0.1. We use Persistent Contrastive Divergence method to sample during the learning process \cite{dl}. We have trained the RBM for 5 epochs with a batch size of 4 and 4 parallel Gibbs Chains, used for sampling using Persistent CD method. Each sentence feature vector is passed through the hidden layer in which feature vector values for each sentence are multiplied by learned weights and a bias value is added to all the feature vector values which is also learned by the RBM. At the end, we have a refined and enhanced matrix. Note that the RBM will have to be trained for each new document that has to be summarized. The idea is that no document can be summarized without going over it. Since each document is unique in the features extracted in section 3.2, the RBM will have to be freshly trained for each new document. 

\subsection{Summary Generation}
The enhanced feature vector values are summed to generate a score against each sentence. The sentences are then sorted according to decreasing score value. The most relevant sentence is the first sentence in this sorted list and is chosen as part of the subset of sentences which will form the summary. Then the next sentence we select is the sentence having highest Jaccard similarity with the first sentence, selected strictly from the top half of the sorted list. This process is recursively and incrementally repeated to select more sentences until a user-specified summary limit is reached. The sentences are then re-arranged in the order of appearance in the original text. This produces a coherent summary rather than a set of haywire sentences.

\section{Results and Performance Evaluation}
Several factual reports from various domains of health, technology, news, sports etc. with varying number of sentences were used for experimentation and evaluation. The proposed algorithm was run on each of those and system-generated summaries were compared to the summaries produced by humans.

\begin{figure}
\centering

\includegraphics[height=6.5cm]{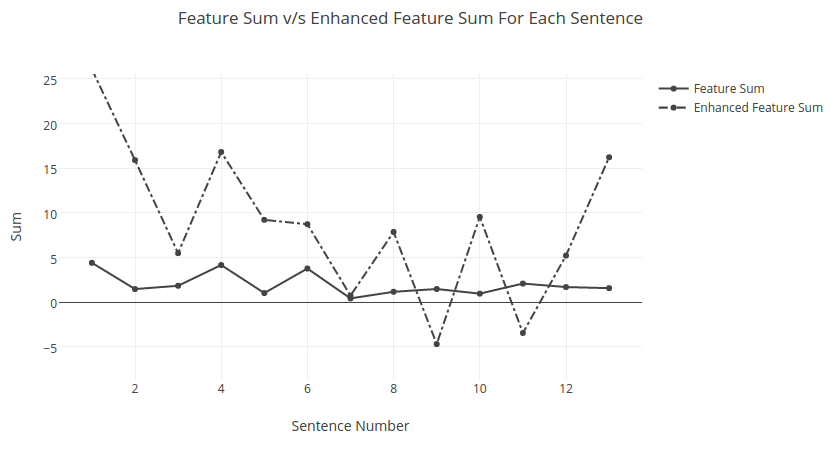}
\centering
\caption{Comparison between feature vector sum and enhanced feature vector sum}

\label{fig:sum}

\end{figure}
Feature Extraction and Enhancement is carried out as proposed in sections 3.2 and 3.3 for all documents. The values of feature vector sum and enhanced feature vector sum for each sentence of one such document have been plotted in Fig 2. The Restricted Boltzmann Machine has extracted a hierarchical representation out of data that initially did not have much variation, hence discovering the latent factors. The sentences have then been ranked on the basis of final feature vector sum and summaries are generated as proposed in section 3.4.
\begin{figure}
\centering

\includegraphics[height=5.7cm]{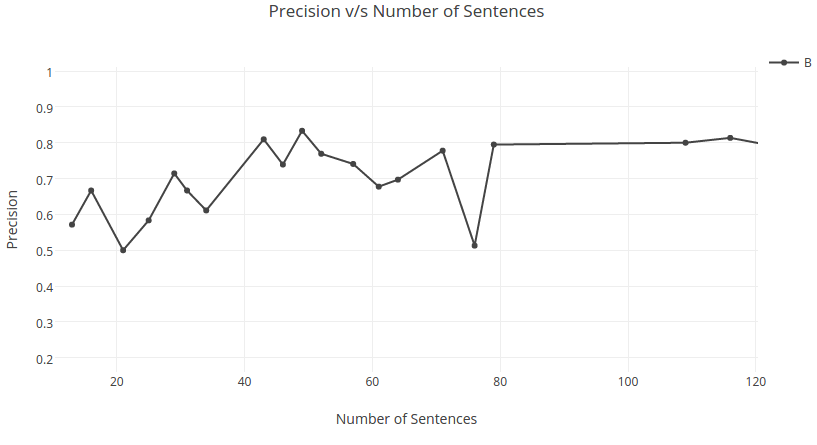}

\caption{Precision values corresponding to summaries of various documents}

\label{fig:p}

\end{figure}

Evaluation of the system-generated summaries is done based on three basic measures: Precision, Recall and F-Measure \cite{ppt}.

\begin{figure}
\centering

\includegraphics[height=5.7cm]{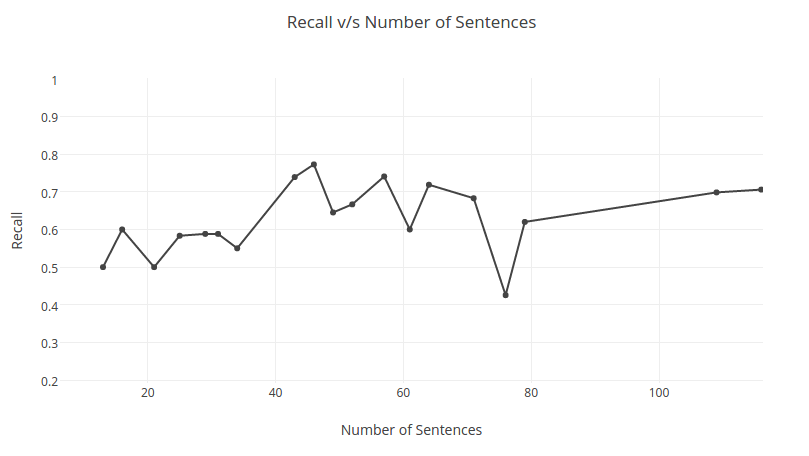}

\caption{Recall values corresponding to summaries of various documents}

\label{fig:r}
\end{figure}

It can be seen that as the number of sentences in the original document cross a certain threshold, the Restricted Boltzmann Machine has ample data to be trained successfully and summaries with high precision and recall values are generated. See Fig 3 and 4.
\begin{figure}
\centering

\includegraphics[height=5.7cm]{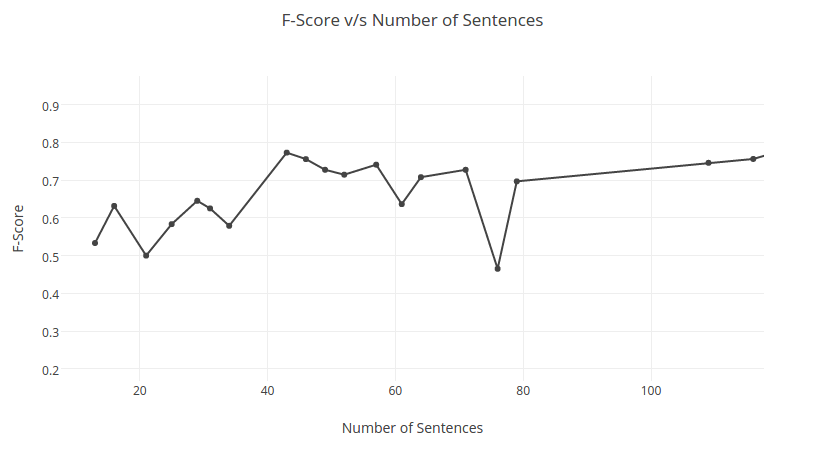}

\caption{F-Measure values corresponding to summaries of various documents}

\label{fig:f}

\end{figure}

F-Measure is defined as follows \cite{2014}:
\begin{equation}
F-Measure = \frac{2*Recall*Precision}{Recall+Precision}
\end{equation}

\section{Comparative Analysis}
The existing approach was executed for the same set of articles with just one layer of RBM, rather than two as it specifies and average values of Precision, Recall and F-Measure were plotted for drawing a comparison between the existing approach and the proposed approach, while keeping the amount of computation  constant. 

\begin{figure}
\centering

\includegraphics[height=5.7cm]{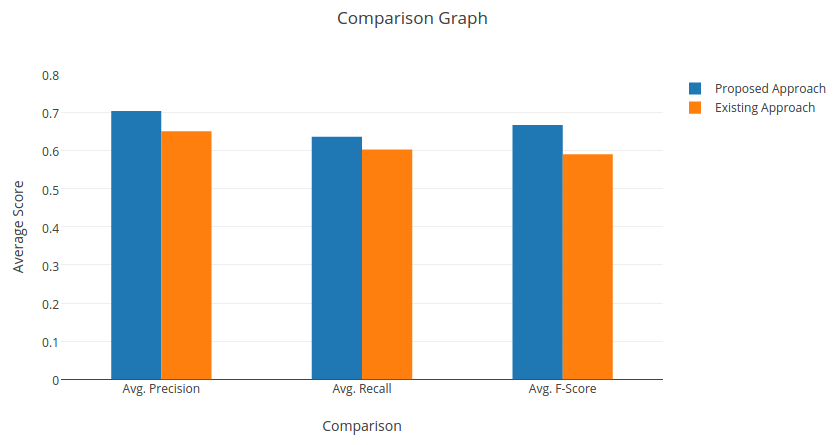}

\caption{Precison, Recall and F-Measure values for the proposed approach (\textit{left bars}) and the existing approach (\textit{right bars})}

\label{fig:comp}

\end{figure}
The proposed approach has an average precision value of 0.7 and average recall value of 0.63 which are both higher than those of the existing approach. Hence, the proposed approach responds better for summarization of factual reports.

\section{Conclusion}
We have developed an algorithm to summarize single-document factual reports. The algorithm runs separately for each input document, instead of learning rules from a corpus, as each document is unique in itself. This is an advantage that our approach provides. We extract 9 features from the given document and enhance them to score each sentence. Recent approaches have been using 2 RBMs stacked on top of each other for feature enhancement. Our approach uses only one RBM and, works effectively and efficiently for factual reports. This has been demonstrated by hand-picking factual descriptions from several domains and comparing the system-generated summaries to those written by humans. This approach can further be developed by adapting the extracted features as per the user's requirements and further adjusting the hyperparameters of the RBM to minimize processing and error in encoded values.  

\subsubsection*{Acknowledgments.}
We would like to extend our gratitude to Dr. Daya Gupta, Professor, Department of Computer Science and Engineering, Delhi Technological University for providing insight and expertise that greatly assisted this research.

\end{document}